\documentclass{article}
\usepackage{spconf,amsmath,graphicx}

\usepackage{amssymb}
\usepackage{enumitem}
\usepackage{epsfig}

\graphicspath{{Figs/}}
\usepackage[]{algorithm2e}

\title{Quickest Intruder Detection for Multiple User Active Authentication}
\name{Pramuditha Perera$^{\star \dagger}$, Julian Fierrez$^{\dagger}$, Vishal M. Patel$^{\star \dagger}$\thanks{Supported by NSF grant 1801435, MINECO/FEDER project BIBECA, and H2020 projects PRIMA, TRESPASS-ETN, and IDEA-FAST. }}
\address{$^{\star \dagger}$ Department of Electrical \& Computer Engineering, Johns Hopkins University \\
	\{pperera3, vpatel36\}@jhu.edu\\
	$^{\dagger}$School of Engineering, Universidad Autonoma de Madrid\\
julian.fierrez@uam.es}

\begin{document}
\ninept
\maketitle
\begin{abstract}
In this paper, we investigate how to detect intruders with low latency for Active Authentication (AA) systems with  multiple-users. We extend the Quickest Change Detection (QCD) framework to the multiple-user case and formulate the Multiple-user Quickest Intruder Detection (MQID) algorithm. Furthermore, we extend the algorithm to the data-efficient scenario where intruder detection is carried out with fewer observation samples. We evaluate the effectiveness of the proposed method on two publicly available AA datasets on the face modality.
\end{abstract}
\section{Introduction}
Balancing the trade-offs between security and usability is one of the major challenges in mobile security \cite{usability_2014}. Longer passwords with a combination of digits, letters and special characters are known to be  secure but they lack usability in the mobile applications. On the other hand, swipe patterns, face verification and fingerprint verification have emerged as popular mobile authentication methods owing to the ease of use they provide. However, security of these methods are challenged due to different types of attack mechanisms employed by intruders ranging from simple shoulder attacks to specifically engineered spoof attacks \cite{ref1}, \cite{ref2}. In this context, Active Authentication (AA), where the mobile device user is continuously monitored and user's identity is continuously verified, has emerged as a promising solution \cite{surveying_mobile_auth_2015}, \cite{VMP_SPM_AA_2016}, \cite{ref3}, \cite{FG_EVT}, \cite{btasoc}.

In our previous work \cite{TIFSQCD}, \cite{QCD_BTAS_2016} we identified three characteristics that are vital to a practical AA system: accuracy, latency and efficiency. However, for AA to be deployed in the real-world, it needs to be equipped with another functionality: transferability. Mobile devices are not private devices that people use in isolation. In practice, it is common for mobile devices to be used interchangeably among several individuals. For example, these individuals could be the members of a family or a set of professionals operating in a team (such as physicians in a hospital). Therefore, it is important that the AA systems facilitate smooth transition between multiple  enrolled individuals \cite{FG_MAA}, \cite{TIFSMAA}.   

\begin{figure}[t]
	\centering
	\includegraphics[width=.48\textwidth]{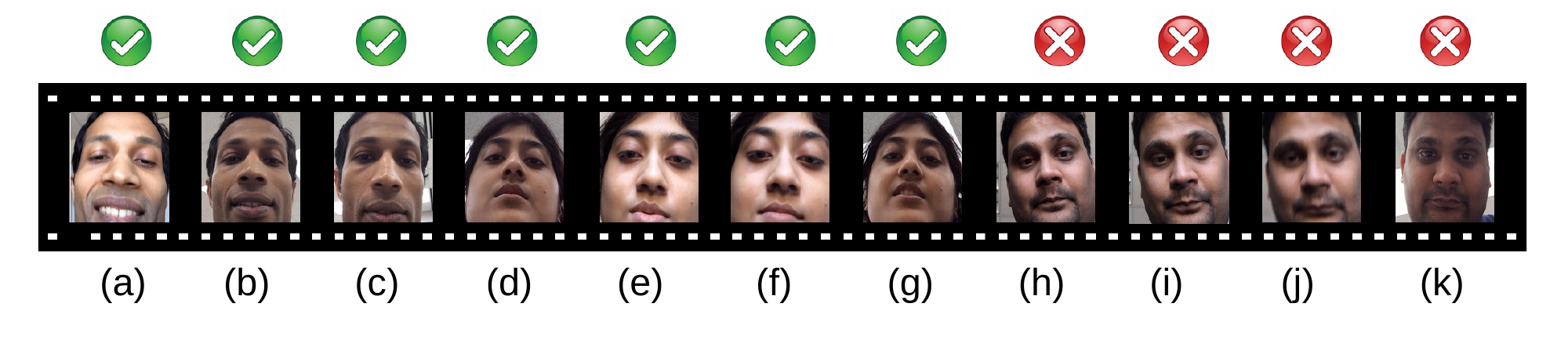} 
	\vskip-15pt\caption{Problem of quickest detection of intruders in multiple-user active authentication. In this example, there are two users enrolled in the mobile device. First user uses the device between frames (a) to (c). At frame (d), another legitimate user starts using the device. The second user uses the device between frames (d) to (g). At frame (h), an intruder starts using the device. The goal of quickest intrusion detection is to detect the change with the lowest possible latency. However, intruder detector should not declare a false detection prior to frame (h).}\label{fig:film}
\end{figure}

The presence of multiple enrolled subjects poses additional challenges to an AA system. Detecting intrusions with low latency in this scenario is even more challenging.  With this new formulation, the device cannot simply declare an intrusion when there is a change in the device usage pattern. This is because two legitimate users operating on the phone could potentially have different behavior patterns \cite{ref4}. As a result, the system is
not only expected to identify intrusions, but also  to provide smooth functioning when there is a transfer of legitimate users. For example, consider the scenario shown in Figure~\ref{fig:film}. There are two legitimate users of the device in this scenario. The first user operates the mobile device between frames (a) and (c). At frame (d), the device is handed over to a second legitimate user. At this point, although there is a change in pattern in device usage, the AA system should not declare an intrusion. On the other hand, when an intruder starts using the device at frame (h), the device is expected to declare an intrusion.

In this paper, we extend our previous work proposed in \cite{TIFSQCD} and study the effectiveness of Quickest Change Detection (QCD) algorithm for multiple-user AA. Specifically, we study possible strategies that can be used to extend Mini-max QCD in AA to the case where multiple users are enrolled in the device. Furthermore, we study the effectiveness of data-efficient sampling for this case. In the experimental results section, we show that QCD algorithm and it's data-efficient extension are effective even in the case of multiple-user AA. 

\section{Proposed Method}

When a user or multiple users start using a mobile device, typically they are required to register with the device. This process in called enrollment of the user(s) to the mobile device. During enrollment, the device gathers sensor observations of the legitimate users and creates user-specific classifiers. Let $U$ be the number of users enrolled in a given device. Technically, $U$ could be any finite number greater or equal to one. However, in practice, it's not common for a mobile device to be shared between more than 5-7 individuals (i.e. normal family size). 

For each user $i$, the device gathers enrollment data $Y_i = \{y_{i,1},y_{i,2},\dots, y_{i,k}\}$. Then, the device develops a set of user specific classifiers $c_i$ for each user which produce classification scores for each user. The classifier $c_i$ can be a simple template matching algorithm or a complex neural network. In our experiments, we consider a template matching algorithm due to the easiness in training the classifier. Our template matching classifier $c_i$ generates a user specific score $s_i = c_i(y) = \min_k(\cos(y,Y_i))$ for a given input $y$ where $\cos(.)$ is the Cosine angle between the two inputs\footnote{Score $s_i$ represents dissimilarity}.

\begin{figure}[t]
	\centering
	\includegraphics[width=.3\textwidth]{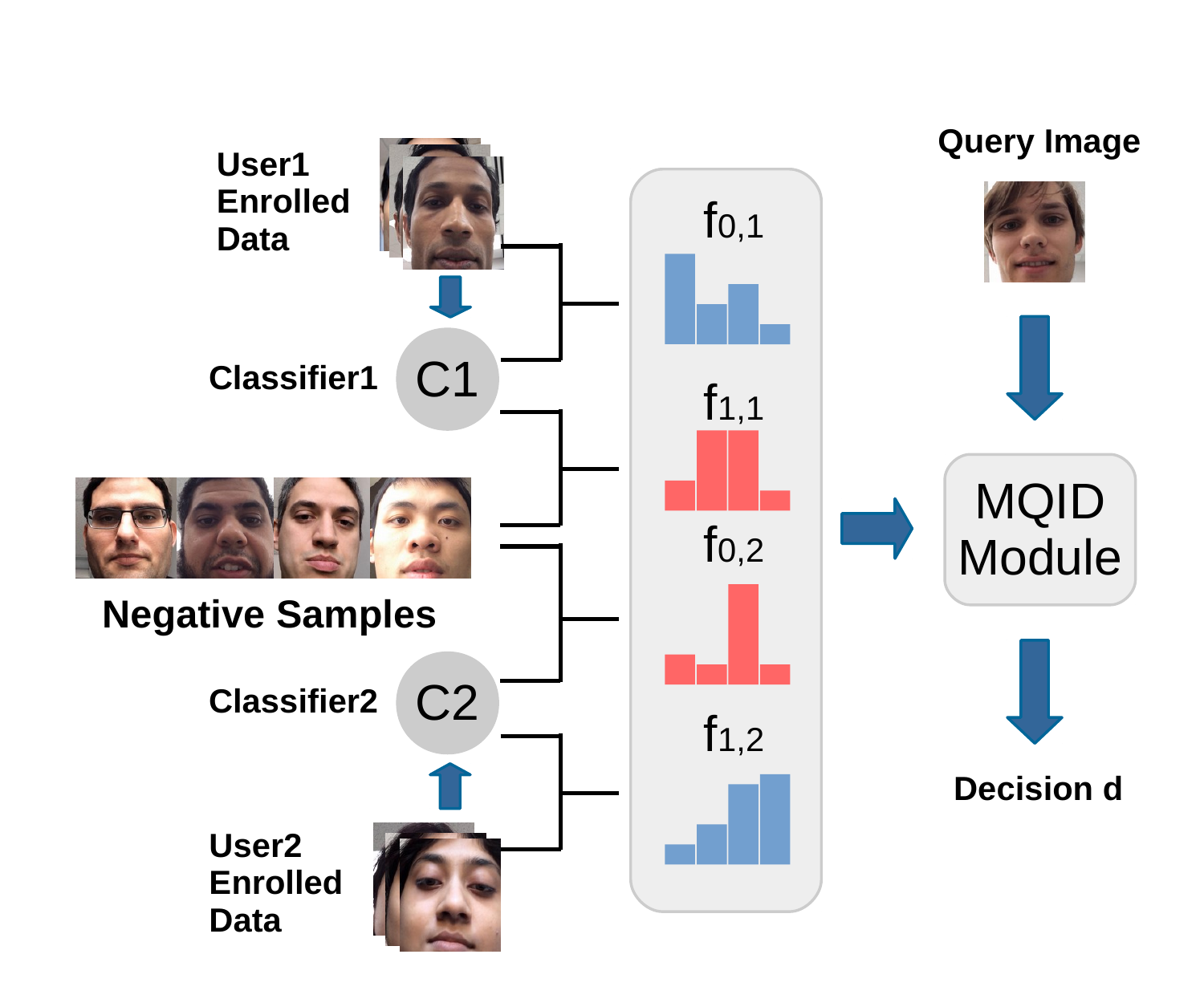} 
\vskip-17pt	\caption{Overview of the problem setup for the case of two enrolled users. For each enrolled user, $i$, the enrolled images are obtained during the enrollment phase. These images are used to train a user specific classifier $c_i$. A matched score distribution $f_{0,i}$ and a non-matched distribution $f_{1,i}$ is obtained for each user.  A known set of negative users are used to obtain the latter. If more users are present the same structure will be cascaded. During inference, Multi-user Quickest Intruder Detection (MQID) module will produce a decision ($d$) by considering the obtained distributions and past decision scores.}\label{fig:over}
\end{figure}

In addition, matched and non-match distributions with respect to the learned classifier are obtained and stored during the enrollment phase. Match distribution $f_{0,i}(.)$ of user $i$  can be obtained by considering pairwise score values of $Y_i$ with respect to $c_i$. On the other hand, a known set of negative samples can be used to obtain the non-matched scores $f_{1,i}(.)$ of user $i$. This process is illustrated in Figure~\ref{fig:over}. In this work, we approximate the score distribution of intruders with the non-matched distribution. Therefore, we use the terms matched distribution and pre-change distribution interchangeably. Similarly, in the context of this paper, non-matched distribution and post-change distribution will also mean the same.  

As the AA system receives observations $\{x_1, x_2, \dots, x_N \}$, and at time $n<N$, it produces a decision $d_n = f(C(x_1),\dots, C(x_n)) \in \{0,1\}$ based on the set of classifiers $C = \{c_1,\dots, c_U\}$ where $f(.)$ is a mapping function. If $d_n = 1$, an intrusion is declared.  Given this formulation, the goal of an AA system is to detect intrusions with the lowest possible latency when a new observation is received. If an intrusion occurs at time $T$, the following two properties are desired from the AA system.

\begin{itemize}[noitemsep]
	\item \textbf{Low detection delay.} The latency between an intrusion occurring and the system detecting the intrusion should be low. If the system detects an intrusion at time $\tau$, the detection delay is given by $(\tau-T)^+$ where $(x)^+$ denotes the positive part of $x$. For all users, this property is quantified using the Average Detection Delay (ADD) defined as  $ADD(\tau) = E[(\tau-T)^+]$.  Here $T$ denotes the real change point. 
	\item  \textbf{Low false detections.} In practice, the detection delay alone cannot characterize the desired functionality of an AA system. It is also desired that the AA system does not produce false detections prior to the intrusion point. These events can be quantified by considering the Probability of False Detections (PFD) as $PFD(\tau) = P[\tau<T]$.  
\end{itemize} 
It is desired for an AA system to have low ADD and low PFD. 

\subsection{Efficient Quickest Change Detection}
Quickest Change Detection (QCD) is a branch of statistical signal processing that thrives to detect the change point of statistical properties of a random process \cite{QCD}, \cite{QCD_MINMAX}, \cite{QCD_explained}. The objective of QCD is to produce algorithms that detect the change with a minimal delay (ADD) while adhering to false alarm rate constraints (PFD). Consider a collection of obtained match scores, $s_1, s_2, \cdots, s_N,$ from the AA system. Assuming that the individual scores are mutually independent, QCD theory can be used to determine whether a change has occurred before time $n$ or not.  

Consider a sequence of time instances $n=1,2,\cdots,N$ in which the device operates.  At each time $n, n>0$, a decision is made whether to take or skip an observation at time $n+1$.  Let $M_{n}$ be the indicator random variable such that $M_{n}=1$ if the score $x_n$ is used for decision making, and $M_{n}=0$ otherwise.  Thus, $M_{n+1}$ is a function of the information available at time $n$, i.e. $M_{n+1} = \phi_{n}(I_n),$ where $\phi_{n}$ is the control law at time $n$, and $I_n =[M_1,M_2,\cdots,M_{n}, s_1^{M_{1}}, s_2^{M_{2}}, \cdots, s_{n}^{M_{n}}]$ represents the information at time $n$. Here, $s_{n}^{M_{n}}$ represents $s_{n}$ if $M_{n}=1$, otherwise $x_{n}$ is absent from the information vector $I_{n}$.  Let $S$ be the stopping time on the information sequence $\{I_{n}\}$.  Then, average percentage of observations (APO) obtained prior to the change point can be quantified as $APO = E\bigg[\frac{1}{S}\sum_{n=1}^{S}M_n\bigg]$, where $E$ denotes the Expected value.

%

In a non-Bayesian setting, due to the absence of a priori distribution on the change point, a different quantity should be used to quantify the number of observations used for decision making. Work in \cite{QCD_MINMAX}, \cite{QCD_explained}, proposes Prechange Duty Cycle (PDC) as $PDC = \limsup_{n} \frac{1}{n} E_n \bigg[ \sum_{k=1}^{n-1} M_k | \tau \geq n \bigg]$
for this purpose. It should be noted that both PDC and APO are similar quantities. With the definition of PDC, efficient QCD in a minimax setting can be formulated as the following optimization problem 
\begin{equation}\label{eq:EMQCD}
\begin{aligned}
& \underset{\phi,\tau}{\text{minimize}}
& & ADD(\phi , \tau) \\
& \text{subject to}
& & PFD(\phi,\tau) \leq \alpha, \;\;PDC(\phi,\tau) \leq \beta. \\
\end{aligned}
\end{equation}     
In \cite{QCD_MINMAX}, a two threshold algorithm called DE-CumSum algorithm, is presented as a solution to this optimization problem. For suitably selected thresholds chosen to meet constraints $\alpha$ and $\beta$, it is shown to obtain the optimal lower bound asymptotically as $\alpha \rightarrow 0$. The DE-CumSum algorithm is presented below.

Start with $W_0 = 0$ and let $\mu >0, A>0 $ and $h \geq 0$. For $n \geq 0$ use the following control rule  $M_{n+1}= 0$ if $W_n < 0$ otherwise $1$ if $W_n \geq 0$.
Statistic $W_n$ is updated as follows
\[W_{n+1}  =
\begin{cases}
\min(W_n+\mu , 0),      & \quad \text{if } M_{n+1} = 0\\
\max(W_n+\log L(s_{n+1}),-h), & \quad \text{if } M_{n+1} = 1,\\
\end{cases}
\]
where $L(s)= \frac{f_1(s)}{f_0(s)}.$  A change is declared at time $\tau_W$, when the statistic $W_n$ passes the threshold $A$ for the first time as $\tau_W = \mathrm{inf}\{n \geq 1 : W_n > A\}.$

\subsection{Multi-user Quickest Intruder Detection (MQID)}  
Based on the discussion above, we introduce the Multiple-user Quickest Intruder Detection (MQID) algorithm. Whether an intrusion has occurred or not is determined using a score value. When the score value is above a pre-determined threshold, an intrusion is declared.  At initialization, it is assumed that the user operating the device is a legitimate user; therefore the score(in our case a dissimilarity score, i.e. a distance) is initialized with zero. The algorithm updates the score value when new observations come. During the update step, the algorithm considers matched and non-matched distributions of all users along with the current score value to produce the updated score. Pseudo code of the algorithm is shown in Algorithm~\ref{alg:DEMQCD}.

The algorithm has three arguments. Argument $\mathit{Efficient}$ determines whether to use data-efficient version of QCD or not. If data-efficient QCD is used then the parameter $\gamma$ determines the floor threshold. Parameter $D$ governs how fast the score is increased. 

During training, enrolled images of each user along with the known negative dataset is used to construct matched and non-matched score distributions. In addition, enrolled images of the user are used to construct a classifier $c_i$. During inference, given an observation $x$, first classification scores from each classifier are obtained. Then, the likelihood of the obtained classifier score is evaluated using the likelihood ratio of each matched and non-matched distribution belonging to each user. The minimum likelihood ratio is considered as the statistic to update the current score of the system.

Updating the score based on the distribution is done as per the Algorithm considering the parameters as well as the magnitude of previous score value.

\IncMargin{1em}
\begin{algorithm}[htp!]
	\SetKwData{Left}{left}\SetKwData{This}{this}\SetKwData{Up}{up}
	\SetKwFunction{Union}{Union}\SetKwFunction{FindCompress}{FindCompress}
	\SetKwInOut{Input}{input}\SetKwInOut{Output}{output}
	\Input{$score,x_n,\{f_{0,i} ,f_{1,i} , c_i  | \forall i \},\gamma,D, \mathit{Efficient}$}
	\Output{$score$}
	\BlankLine
	
	$L = \min_i \log(\frac{f_{1,i}(c_i(x_n))}{f_{0,i}(c_i(x_n))})  $
	
	\uIf{$\mathit{Efficient}$}{
		\uIf{$score<0$}{{$score =\min(\mathit{score}+D,0) $\;}}
		\Else{$score \leftarrow \max(score+L, -\gamma)$ 	\;}
	}
	\Else{$score \leftarrow score+L$ 	\;}
	
	Return ($score$)\;
	
	\caption{Algorithm to update the score based on the observations for the proposed method. $n$ denotes time and $i$ denotes individual.}\label{alg:DEMQCD}
\end{algorithm}\DecMargin{1em}

\section{Experimental Results}
\begin{figure}[t!]
	\centering
	\includegraphics[width=.35\textwidth]{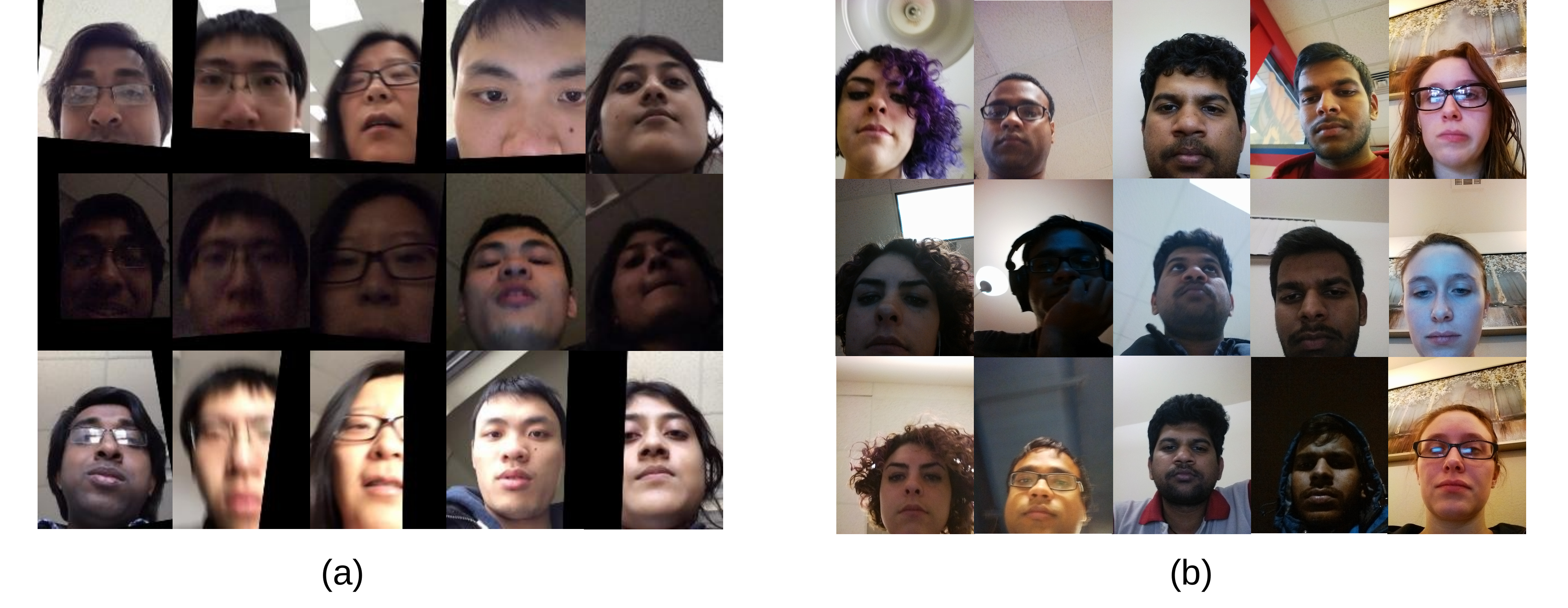} 
	\vskip -15pt\caption{Sample face images from the (a) UMDAA01 dataset and (b) UMDAA02 dataset used for evaluation. Samples from the same subject are shown in each column.}\label{fig:dataset}
\end{figure}

We test the proposed method on two publicly available Active Authentication datasets - UMDAA01 \cite{UMDAA} and UMDAA02 \cite{UMDAA02}  using the face modality. The UMDAA-01 dataset \cite{UMDAA} contains images captured using the front-facing camera of an iPhone 5S mobile device of 50 different individuals captured across three sessions with varying illumination conditions, see Figure~\ref{fig:dataset}(a). 


The UMDAA-02 Dataset \cite{UMDAA02} is an unconstrained multimodal dataset with 44 subjects where 18 sensor observations were recorded across a two month period using a Nexus 5 mobile device. Authors of \cite{UMDAA02} have made the face modality and the touch-data modality \cite{2018_TIFS_Swipe_Fierrez} publicly available. In our work we only consider the face modality to perform tests. A sample set of images obtained from this dataset is shown in Figure~\ref{fig:dataset}(b). 



\noindent {\bf{Protocol: }}
In both datasets, the first 22 users were used as possible enrolled users. Users 23-33 were used as the known negative samples. Remaining users were considered as intruders. From the enrolled users $10\%$ of data were randomly chosen to represent the enrolled images. These image frames were removed from the test set. For each dataset, we varied the number of enrolled users from 1 to 7. If the number of enrolled users is $U$, we partitioned the first 22 users into disjoint groups of $U$ and carried out $\mathrm{floor}(22/U)$ trials. 


In order to simulate an intrusion, the following process was followed. The entire video clips of the enrolled users were appended in the order of their index to form an augmented video for each trial. Then, each intruder from the intruder set was considered one at a time. Considered intruder's video clip was appended at the end of the augmented video clip to produce the test video clip. Shown in Figure~\ref{fig:film} is a summary of such a clip for the case of two enrolled users.  During training, we extracted the image frames from the video clip with a sampling rate of 1 image per 3 seconds. We used the Viola-Jones face detector to detect faces in the extracted image frame and performed local histogram normalization. The extracted image was resized to $224\times 224$ and image features were extracted from the ResNet18 deep architecture which was pre-trained on the ImageNet dataset. For all cases, we considered the distance to the nearest neighbor as the user specific classifier $c_i$.

\begin{table*}[htp!]
	\centering
	\caption{Tabulation of ADD for PFD of 2\% and 5\% when the users are varied from 1 to 7 on the UMDAA01 dataset. When a particular method failed to achieve prescribed PFD it is indicated by N/A.}
	\label{table:umd1}
	\resizebox{.75\linewidth}{!}{
		\begin{tabular}{|l|l|l|l|l|l|l|l|l|l|l|l|l|l|l|}
			\hline
			\textbf{}             & \multicolumn{7}{c|}{\textbf{5\%}}                                                        & \multicolumn{7}{c|}{\textbf{2\%}}                                                        \\ \hline
			\textbf{\# of Users}  & \textbf{1} & \textbf{2} & \textbf{3} & \textbf{4} & \textbf{5} & \textbf{6} & \textbf{7} & \textbf{1} & \textbf{2} & \textbf{3} & \textbf{4} & \textbf{5} & \textbf{6} & \textbf{7} \\ \hline
			\textbf{SSH}          & 1.14       & 1.14       & 1.17       & 1.18       & 1.25       & 1.35       & 1.31       & 2.28       & 2.49       & 4.73       & 3.98       & 4.41       & 5.84       & 4.49       \\ \hline
			\textbf{Sui et al}    & N/A         & 1.82       & 1.66       & 1.98       & 1.89       & 1.86       & 2.51       & N/A          & 2.04       & 4.74       & 7.74       & 13.41      & 8.59       & 13.41      \\ \hline
			\textbf{Crouse et al} & N/A         & N/A          & 57.20      & 33.92      & 29.65      & 19.49      & 47.8       & 28.4       & N/A          & N/A         & 40.44      & 32.25      & 37.04      & 56.2       \\ \hline
			\textbf{Pn (FG17)}    & 2.10       & 2.20       & 2.16       & 3.29       & 2.35       & 2.23       & 2.51       & 3.96       & 3.84       & 2.89       & 5.48       & 4.41       & 5.06       & 4.51       \\ \hline
			\textbf{MQID (present work)}         & 1.14       & 1.14       & 1.17       & 1.20       & 1.25       & 1.19       & 1.31       & 1.63       & 1.65       & 1.79       & 2.08       & 2.02       & 2.12       & 2.49       \\ \hline
			\textbf{DEMQID (present work)}       & 2.14       & 1.52       & 2.37       & 1.59       & 2.51       & 1.86       & 2.51       & 2.28       & 1.64       & 2.51       & 1.84       & 2.72       & 1.92       & 2.49       \\ \hline
	\end{tabular}}
\end{table*}

\begin{table*}[htp!]
	\centering
	\caption{Tabulation of ADD for PFD of 2\% and 5\% when the users are varied from 1 to 7 on the UMDAA02 dataset. When a particular method failed to achieve prescribed PFD it is indicated by N/A.}
	\label{table:umd2}
	\resizebox{.75\linewidth}{!}{
		
		\begin{tabular}{|l|l|l|l|l|l|l|l|l|l|l|l|l|l|l|}
			\hline
			\textbf{}             & \multicolumn{7}{c|}{\textbf{5\%}}                                                        & \multicolumn{7}{c|}{\textbf{2\%}}                                                        \\ \hline
			\textbf{\# of Users}  & \textbf{1} & \textbf{2} & \textbf{3} & \textbf{4} & \textbf{5} & \textbf{6} & \textbf{7} & \textbf{1} & \textbf{2} & \textbf{3} & \textbf{4} & \textbf{5} & \textbf{6} & \textbf{7} \\ \hline
			\textbf{SSH}          & 19.7       & 19.6       & 51.82      & 24.33      & 52.64      & 10.01      & 110.2      & 72.92      & 90.71      & 150.0      & 36.23      & 107.6      & 93.6       & 118.1      \\ \hline
			\textbf{Sui et al}    & 63.1       & 284.9      & N/A          & 243.1      & N/A          & 237.6      & 109.4      & N/A          & N/A          & N/A          & N/A          & N/A          & N/A          & N/A          \\ \hline
			\textbf{Crouse et al} & 364.5      & N/A          & N/A          & N/A          & N/A          & N/A         & N/A          & 467.8      & N/A          & N/A          & N/A          & N/A          & N/A         & N/A          \\ \hline
			\textbf{Pn (FG17)}    & 2.71       & 3.31       & 10.91      & 11.22      & 34.64      & 7.06       & 44.55      & 4.30       & 39.56      & 72.52      & 37.44      & 86.3       & 64.0       & 116.0      \\ \hline
			\textbf{MQID (present work)}         & 3.83       & 4.28       & 5.42       & 6.67       & 6.11       & 5.61       & 5.30       & 5.58       & 5.77       & 8.14       & 10.38      & 9.10       & 8.03       & 7.78       \\ \hline
			\textbf{DEMQID (present work)}       & 3.47       & 3.17       & 4.13       & 4.618      & 6.93       & 3.85       & 5.82       & 4.32       & 4.39       & 6.38       & 8.34       & 9.12       & 4.79       & 10.15      \\ \hline
	\end{tabular}}
\end{table*}

\noindent {\bf{Metrics: }}
The performance of a quickest change detection scheme depends on ADD and PFD. Ideally, an AA system should be able to operate with low  ADD and a low PFD. In order to evaluate performance of the system following \cite{TIFSQCD}, we used the ADD-PFD graph. 

\noindent {\bf{Methods: }}
We evaluated the following methods using the protocol presented: Single Score-based Authentication (SSA), Time decay fusion (Sui et al.) \cite{two_sample_method}, Confidence functions (Crouse et al.) \cite{Jain_AA_ICB2015}, Probability of Negativity $P_n(FG17)$ \cite{FG_MAA}, Multi-user Quickest Intruder Detection (MQID) -- the method proposed in this paper with the Min-Max formulation \cite{TIFSQCD} and, Data Efficient Multi-user Quickest Intruder Detection (DEMQID)-- The method proposed in this paper using the Min-Max formulation with data-efficient constraints.  For a fair comparison, in all cases except for $P_n(FG17)$ \cite{FG_MAA} we used the statistic $L = \min_i \log(\frac{f_{1,i}(c_i(x_n))}{f_{0,i}(c_i(x_n))})$ as the score value to perform intrusion detection.

\begin{figure}[htp!]
	\centering
		\includegraphics[width=3.5cm]{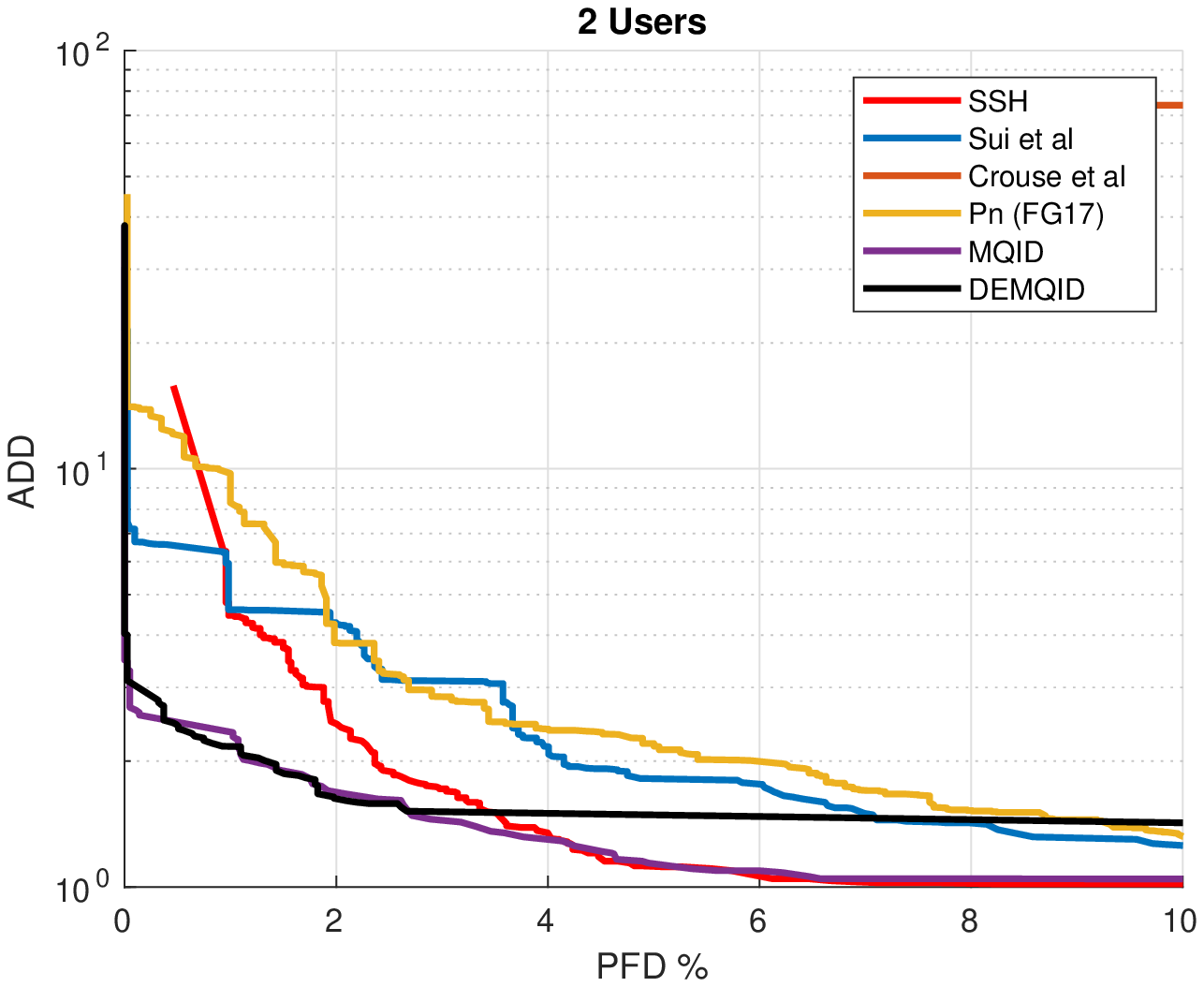}
		\includegraphics[width=3.5cm]{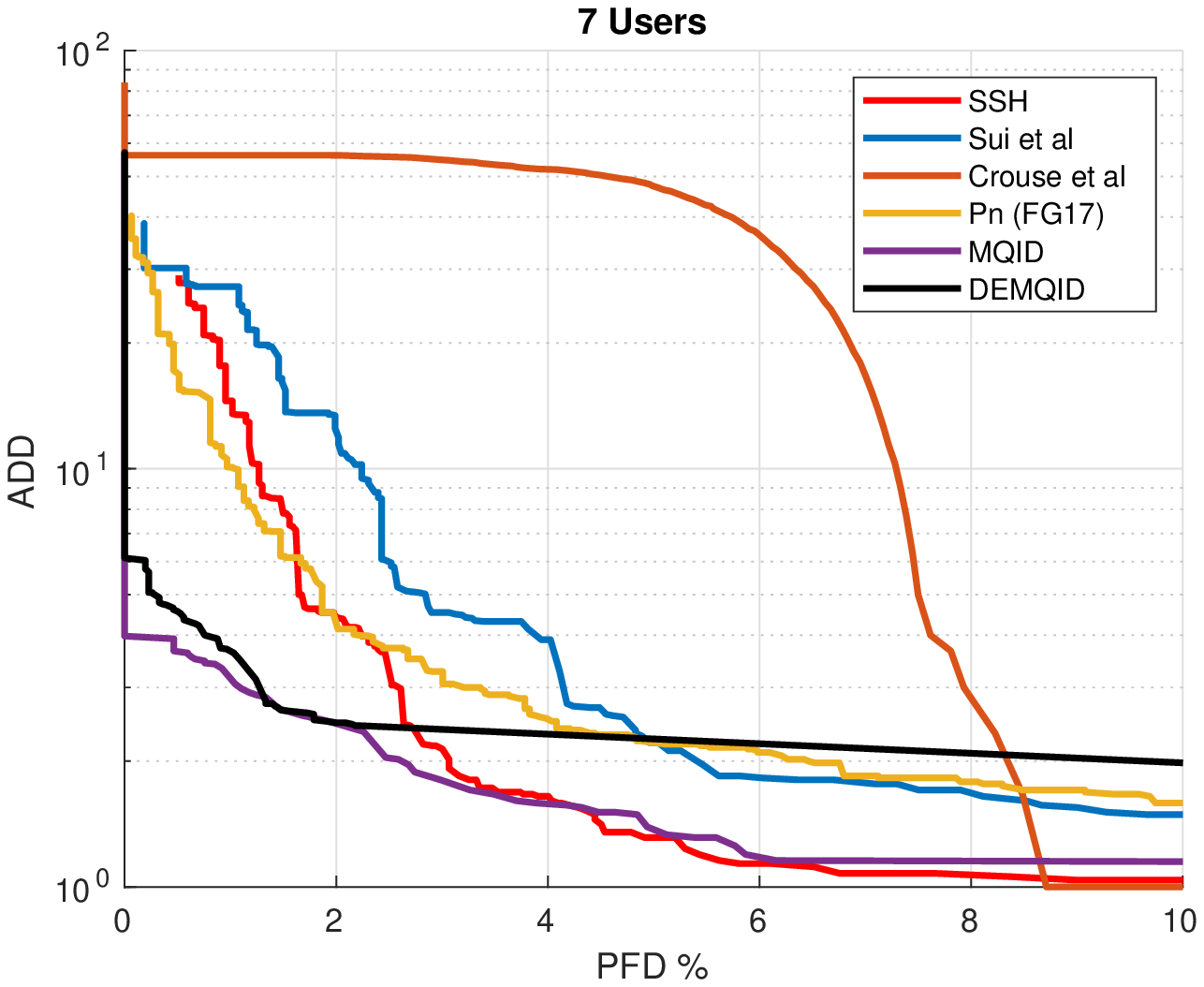}\\
		\includegraphics[width=3.5cm]{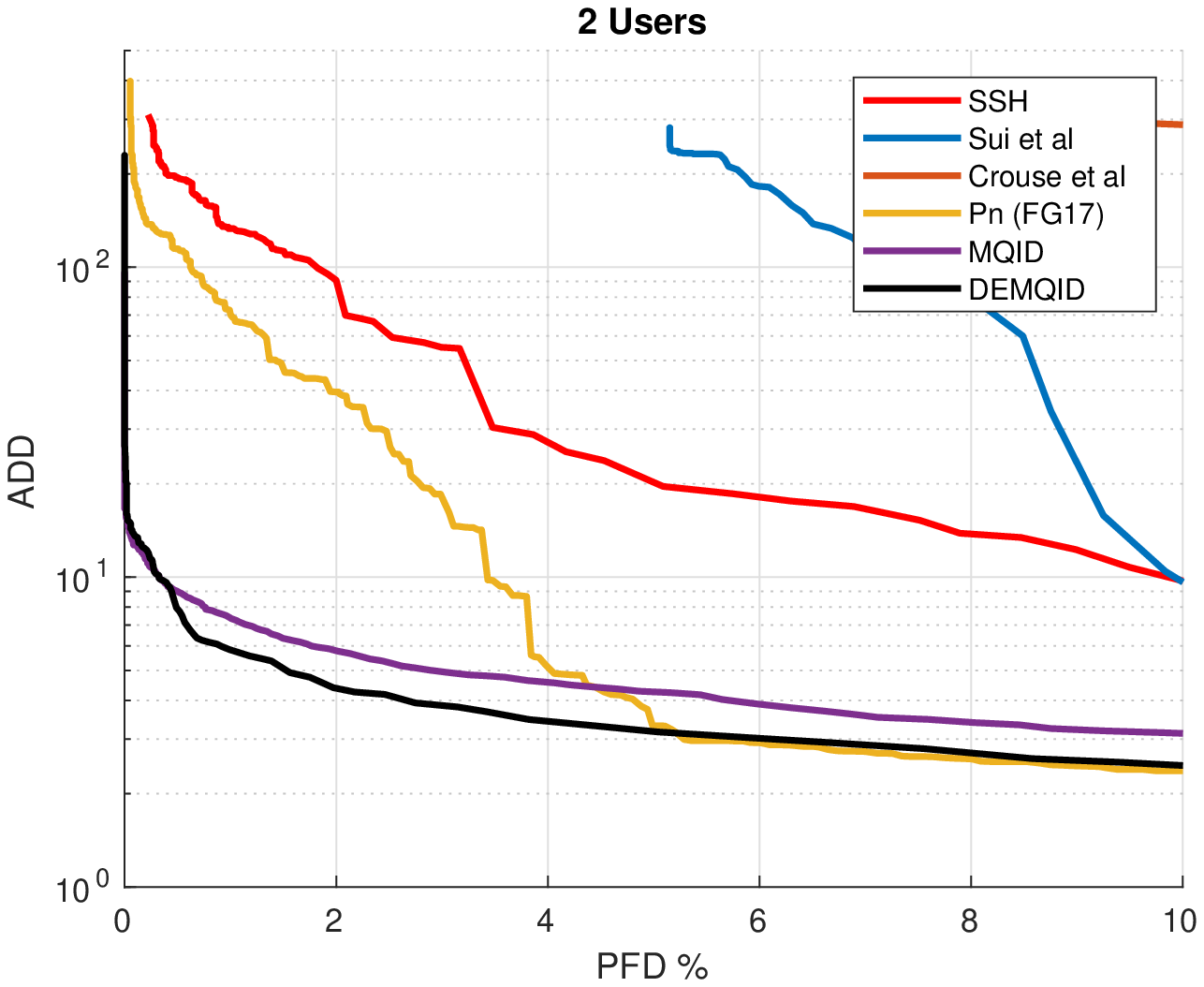}
		\includegraphics[width=3.5cm]{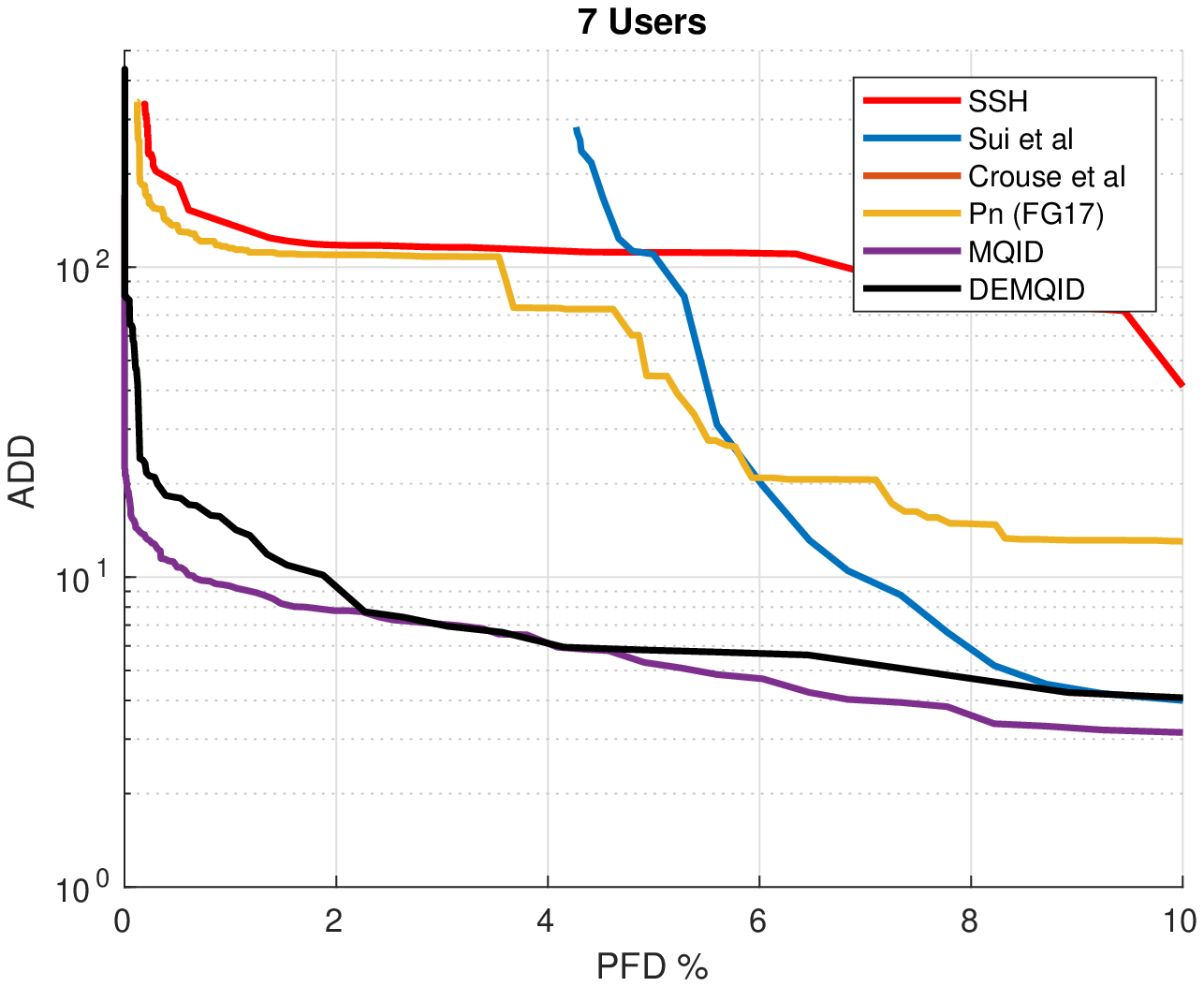}	 
\vskip -10pt	\caption{The ADD-PFD curves corresponding to the UMDAA01 (top row) and UMDAA02 (bottom row) datasets when the number of users are varied from 1 to 7. Due to space limitations only curves corresponding to users 2 and 7 are shown (color image).}\label{umd1Umd2}
\end{figure}

\noindent {\bf{Results: }}
The ADD-PFD curves corresponding to the experiments on the UMDAA01 and UMDAA02 datasets are shown in Figure~\ref{umd1Umd2} when the number of users are varied from 1 to 7. Due to space limitations only curves corresponding to users 2 and 7 are shown.  ADD values obtained for PFD of 2\% and 5\% are tabulated for UMDAA01 and UMDAA02 in Tables~\ref{table:umd1} and Tables~\ref{table:umd2}, respectively. These tables indicate the latency of detecting an intrusion in average while guaranteeing a fixed false detection rate.


\noindent {\bf{Results on UMDAA01: }}
In all considered cases MQID method has performed better than the other baseline methods when it was desired to achieve a PFD of 2\%. It is also seen that $P_n(FG17)$, which is a method proposed for multi-user AA has also outperformed SSH method which uses log-likelihood ratio in all cases. Furthermore, data-efficient version of the algorithm, DEMQID, has performed on par with MQID, even performing better in certain cases. Average percentage of observations obtained in DEMQID for this dataset was 0.304.

However, it can be observed that when 5\% of PFD is allowed, even other baseline methods perform reasonably well. For example, in majority of the cases SSH has performed on par with MQID. We also observe that DEMQID is slightly worse than MQID in this case. This suggests that for the employed deep feature, a PFD rate of 5\% can be obtained even when the sequence of data are not considered.  DEMQID takes more sparse samples when deciding the score value. As a result, when the score function is noisy, DEMQID is not affected by the noise as much as MQID. Even-though sparser sampling would result in some latency in detection, overall trade-off can be beneficial.  This is why DEMQID outperforms MQID when decision making is more challenging (as was the case when PFD of 2\% was considered). 

However, when the decision making becomes easier, DEMQID does not contribute towards improving the detection accuracy as score values are less noisy. This is why in the case of 5\% of PFD, DEMQID performs worse than MQID.


\noindent {\bf{Results on UMDAA02: }}
As a result of having higher complexity, detecting intruders become more challenging in UMDAA02 compared to UMDAA01. However, due the challenging behavior of the dataset, the importance of the proposed method is magnified. In all ADD-PFD curves obtained for UMDAA02 in Figure~\ref{umd1Umd2}, it is evident that the proposed methods significantly outperform the baseline methods. Furthermore, DEMQID has outperformed MQID in most of the cases showing the significance of data efficient QCD.  

In our evaluations we show that even when the number of users are increased, the performance of the proposed system does not drop drastically. For the UMDAA01 dataset, only 2.35 additional samples were required to maintain a probability of false detection of 2\% when the users were increased from 1 o 7. In a more challenging UMDAA02 dataset, 4.33 more samples were required on average to maintain the same false detection rate.

%

	

\section{Conclusion}
It has been previously shown that AA yields superior detection performance when the QCD algorithm is used \cite{TIFSQCD}. However, this is the first work that studies the problem of QCD in a multiple-user AA scenario.   We proposed MQID algorithm for multiple-user AA with low latency. Furthermore, we extended the initial formulation to a data efficient version by proposing DEMQID algorithm. We evaluated the performance of the proposed methods on two face-based AA datasets. Our experiments suggest that the proposed method is more effective compared to the baseline methods we considered. The proposed method allows the number of enrolled users to be increased with a relatively smaller cost in terms of observations. Only 2.35 and 4.33 observations were required on average to maintain a false detection rate of 2\% when the users were increased from 1 to 7 in the UMDAA01 and UMDAA02 datasets, respectively.

\bibliographystyle{IEEEbib}
\bibliography{AA_refs}

\end{document}